\documentclass[journal]{IEEEtran}

\usepackage{amsmath,amssymb,amsfonts}
\usepackage{graphicx}
\usepackage{textcomp}
\usepackage{subfig}
\usepackage{xcolor}
\usepackage{caption}
\usepackage{cite}
\usepackage{listings,xcolor}
\usepackage[utf8]{inputenc}
\usepackage{float}
\usepackage{adjustbox}
\usepackage{etoolbox}
\usepackage{pifont}

\usepackage{framed}
\usepackage{url}
\usepackage{pbox}
\usepackage{flushend}

\newcommand\blfootnote[1]{%
  \begingroup
  \renewcommand\thefootnote{}\footnote{#1}%
  \addtocounter{footnote}{-1}%
  \endgroup
}

\usepackage[hidelinks]{hyperref}
\hypersetup{hidelinks}

\begin{document}

\title{IGNNITION: Bridging the Gap Between\\Graph Neural Networks and Networking Systems}

\author{
\IEEEauthorblockN{David Pujol-Perich,
José Suárez-Varela,
Miquel Ferriol,
Shihan Xiao,\\
Bo Wu,
Albert Cabellos-Aparicio and
Pere Barlet-Ros
\thanks{
$\copyright$2021 IEEE. Personal use of this material is permitted. Permission from IEEE must be obtained for all other uses, in any current or future media, including reprinting/republishing this material for advertising or promotional purposes, creating new collective works, for resale or redistribution to servers or lists, or reuse of any copyrighted component of this work in other works.}
}\\
\vspace{0.3cm}
\small{\textbf{\underline{NOTE:}} Please use the following reference to cite this work:\\ D. Pujol-Perich, J. Suárez-Varela, M. Ferriol, S. Xiao, B. Wu, A. Cabellos-Aparicio, and P. Barlet-Ros, “IGNNITION: Bridging the Gap between Graph Neural Networks and Networking Systems”, \textit{IEEE Network}, vol. 35, no. 6, pp. 171–177, 2021.}
\vspace{-0.9cm}
}

\maketitle
\begin{abstract}
Recent years have seen the vast potential of \textit{Graph Neural Networks} (GNN) in many fields where data is structured as graphs (e.g., chemistry, recommender systems). In particular, GNNs are becoming increasingly popular in the field of networking, as graphs are intrinsically present at many levels (e.g., topology, routing). The main novelty of GNNs is their ability to generalize to other networks unseen during training, which is an essential feature for developing practical Machine Learning (ML) solutions for networking. However, implementing a functional GNN prototype is currently a cumbersome task that requires strong skills in neural network programming. This poses an important barrier to network engineers that often do not have the necessary ML expertise. In this article, we present IGNNITION, a novel open-source framework that enables fast prototyping of GNNs for networking systems. IGNNITION is based on an intuitive high-level abstraction that hides the complexity behind GNNs, while still offering great flexibility to build custom GNN architectures. To showcase the versatility and performance of this framework, we implement two state-of-the-art GNN models applied to different networking use cases. Our results show that the GNN models produced by IGNNITION are equivalent in terms of accuracy and performance to their native implementations in TensorFlow.
\end{abstract}

\maketitle

\blfootnote{\textbf{David Pujol-Perich, José Suárez-Varela, Miquel Ferriol-Galmés, Albert Cabellos-Aparicio, and Pere Barlet-Ros:} Barcelona Neural Networking Center, Universitat Politècnica de Catalunya.\\
\textbf{Shihan Xiao and Bo Wu:} Network Technology Lab., Huawei Technologies Co.,Ltd.}

\vspace{-0.4cm}
\section*{Introduction} \label{Introduction}
\textit{Graph Neural Networks} (GNN)~\cite{battaglia2018relational} have recently become a hot topic among the Machine Learning (ML) community. The main novelty behind these models is their unique ability to learn and generalize over graph-structured information. This has enabled the development of groundbreaking applications in different fields where data is fundamentally represented as graphs (e.g., chemistry, physics, biology, recommender systems, social networks)~\cite{ZHOU202057}.

The last few years have seen an increasing interest among the networking community to exploit the potential of GNN, as many fundamental networking problems involve the use of graphs (e.g., topology, routing). As such, we have already witnessed some successful GNN-based applications for networking (e.g., routing optimization~\cite{rusek2019unveiling, geyer2018learning}, virtual network function placement~\cite{quang2019deep}, resource allocation in wireless networks~\cite{eisen2020optimal}, job scheduling in data center networks~\cite{mao2019learning}).
In particular, GNN is the only ML technique that demonstrated the capability to effectively operate in other networks unseen during the training phase. This is thanks to the generalization power of GNNs over graphs which, in the networking field, represents a cornerstone for the development of a new generation of ML-based solutions that can be trained in a controlled network testbed (e.g., at the vendor's lab) and, once trained, be ready for deployment in any customer network (topologies of arbitrary size and structure), without the need for re-training on premises~\cite{rusek2019unveiling}.

Nowadays, designing and implementing a GNN model involves dealing with complex mathematical formulation and using ML libraries with tensor-based programming, such as \mbox{TensorFlow}, or PyTorch. At the same time, \textit{GNN models} for networking often rely on non-standard GNN architectures (e.g., heterogeneous graphs with sequential dependencies between different element types~\cite{rusek2019unveiling, geyer2018learning}), which makes their implementation even more difficult. This represents a critical entry barrier for network researchers and engineers that could benefit from the use of GNN, but lack the necessary ML expertise to implement these models.

\begin{figure}[!t]
    \vspace{0.1cm}
    \centering
    \includegraphics[width=\columnwidth]{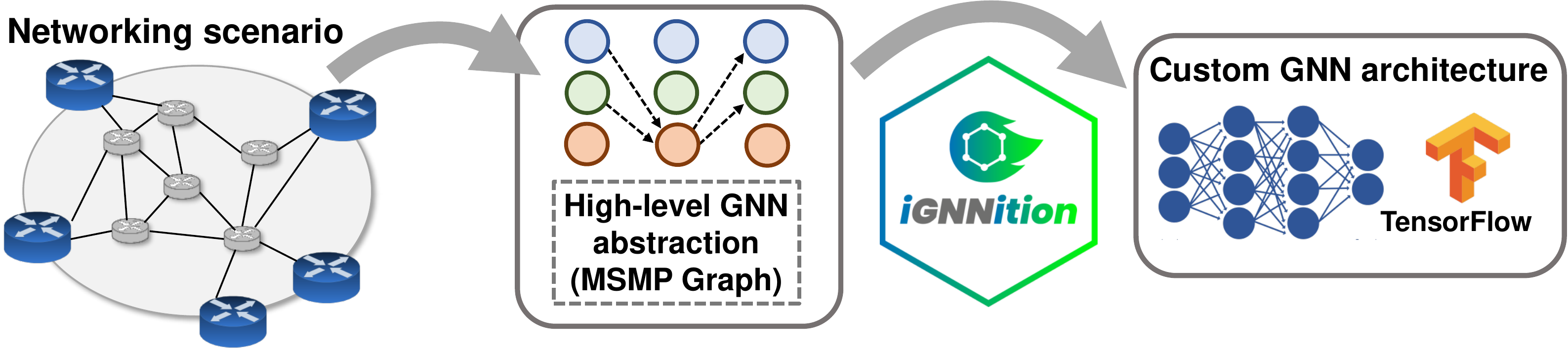}
    \caption{Overview of IGNNITION.}
    \label{fig:overview_ignnition}
\end{figure}
\setlength{\textfloatsep}{0.3cm}

In this article, we present IGNNITION, a TensorFlow-based framework for fast prototyping of GNNs (see Fig.~\ref{fig:overview_ignnition}). This framework is open source\footnote{Available at: \url{https://ignnition.net}}, and mainly targets networking experts and researchers with little background on neural network programming. With IGNNITION, users can easily design their own GNN models \mbox{--- even} for the most complex network \mbox{scenarios ---} via an intuitive, human-readable YAML file. Based on this input, the framework automatically generates an efficient TensorFlow implementation of the GNN, without the need for writing a single line of \mbox{TensorFlow} code.

To achieve this, we propose a high-level abstraction we call the Multi-Stage Message Passing graph (\textit{MSMP graph}). This novel abstraction covers a broad definition of GNN, offering great flexibility to design a wide variety of existing GNN variants, and generate new custom architectures combining individual components of them (e.g., message-passing functions). Likewise, MSMP graphs enable to hide the complex mathematical formulation and the tensor-wise operations behind the implementation of GNNs.

IGNNITION provides a YAML-based \textit{codeless} interface that network engineers and practitioners can leverage to implement custom GNN models applied to networking. In this context, coding these types of models with general-purpose GNN libraries (e.g.,~\cite{you2020design,battaglia2018relational, fey2019fast, wang2019dgl,grattarola2020graph, ma2019neugraph}) is arguably more complex for non experts in neural network programming, as they often require using lower-level abstractions, and some of these libraries even lack sufficient flexibility to implement non-standard GNN architectures applied to networking~\cite{you2020design,grattarola2020graph}.

More in detail, IGNNITION offers the following main features:
\begin{itemize}
    \item \textbf{High-level abstraction:} Codeless interface \mbox{--- based} on MSMP graphs and YAML files --- that abstracts away both the mathematical formulation of the GNN and its implementation.
    \item \textbf{Flexibility of design:} Broad support for existing GNN variants and custom message-passing architectures via the novel MSMP graph abstraction.
    \item \textbf{High performance:} Efficient GNN implementations equivalent to native TensorFlow code, as performance is crucial for network applications.
    \item \textbf{Easy debugging:} Advanced error-checking mechanisms and enhanced visualizations of the implemented GNNs, to help network engineers troubleshoot their models.
    \item \textbf{Easy integration:} User-friendly dataset interface, based on the well-known NetworkX library, to feed GNN models with data from different sources and in various formats (e.g., network monitoring logs).
    \item \textbf{Networking focus:} IGNNITION is specially designed to facilitate the implementation of non-standard features commonly used in GNN-based networking applications (e.g., support for heterogeneous graphs with possible sequential dependencies~\cite{rusek2019unveiling, geyer2018learning}).
\end{itemize}

As a result, with IGNNITION, network researchers and engineers with limited knowledge on neural network programming should be able to quickly produce a fully functional prototype of a GNN model tailored to a particular application.

\section*{Background on GNN} \label{Graph Neural Networks} 

The input of a GNN is a graph, which comprises a set of nodes and edges connecting them. Each node $v$ has an associated hidden state vector $h_\text{v}$ of predefined size that encodes its internal state. At the beginning of the GNN execution, hidden state vectors are initialized with some node-related features included in the input graph. 

\begin{figure}[!t]
    \centering
    \vspace{0.1cm}
    \includegraphics[width=\columnwidth]{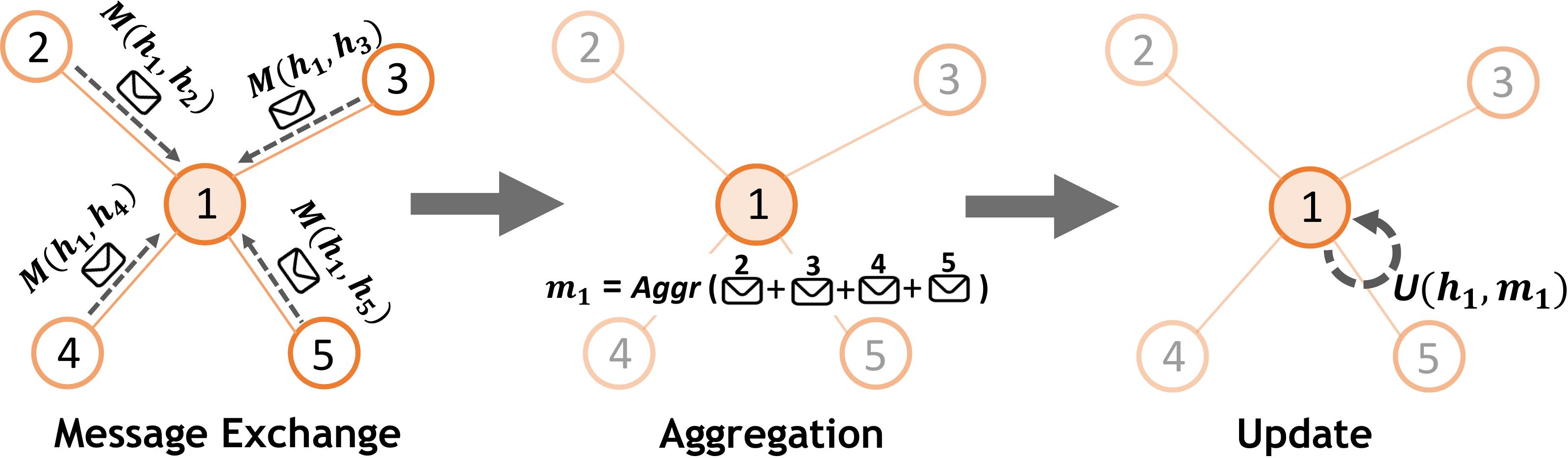}
    \caption{Phases of a message-passing iteration: Message exchange, Aggregation, and Update}
    \label{fig:background}
\end{figure}

After this, a message-passing algorithm is executed according to the connections of the input graph. This message-passing process comprises three main phases (see Fig.~\ref{fig:background}): 
\begin{enumerate}
\item Message exchange
\item Aggregation
\item Update
\end{enumerate}

First, each node sends a message to all its neighbors in the graph. This message is the result of combining the hidden states of the sender node $h_v$ and the receiver neighbor $h_u$ through the so-called \textit{Message} function ($M$). After this, each node aggregates all the messages collected from its neighbors into a single fixed-size vector $m_\text{v}$ (a.k.a., the aggregated message). To do this, nodes use a common \textit{Aggregation} function ($Aggr$), which in standard GNNs is typically an element-wise summation over all the messages collected on the node. Lastly, each node applies an \textit{Update} function ($U$) that combines its own hidden state $h_\text{v}$ with the node's aggregated message $m_\text{v}$. This message passing is repeated a number of iterations $T$, until the nodes' hidden states $h_v$ converge to some fixed values.

\setlength{\textfloatsep}{0.3cm}

After the message-passing phase, a \textit{Readout} function is used to produce the output of the GNN model. This function takes as input the nodes' hidden states and converts them into the final predictions of the model. Note that the output of the Readout function can be either a set of global graph-level properties or specific node-level features.

An essential aspect of GNN is that the internal functions that shape its architecture are modeled by neural networks (NN). Particularly, the \textit{Message}, \textit{Update}, and \textit{Readout} functions are typically approximated by three independent NNs (e.g., feed-forward NN, Recurrent NN). These NNs are then dynamically assembled within the GNN architecture according to the nodes and connections of the input graphs. During training, the aforementioned NNs (e.g., the \textit{Message} NN) optimize jointly their internal parameters, effectively learning the target function from the perspective of each individual node in the graph. As a result, after training, the GNN is able to make accurate predictions on graphs with different sizes and structures not seen before. 

Note that beyond the generic GNN description provided in this section, there is a rich body of literature with different GNN architectural variants (e.g., Graph Convolutional Networks, Gated Neural Networks, Graph Attention Networks, Graph Recurrent Networks, GraphSAGE)~\cite{ZHOU202057}, while all of them share the basic principle of an initial message-passing phase followed by a final readout. We refer the reader to~\cite{battaglia2018relational, ZHOU202057} for more generic background on GNN.

\vspace{-0.2cm}
\section*{The MSMP Graph Abstraction} \label{Sec: Graph abstraction}

In the networking field, GNN models need to be highly customized to accommodate specific networking use cases. Particularly, it is common to find non-standard message-passing architectures~\cite{rusek2019unveiling,geyer2018learning, eisen2020optimal} that stem from the need to explicitly model the circular dependencies between various network elements, based on the actual behavior of real network infrastructures. As a result, state-of-the-art GNN models for networking often consider heterogeneous graphs as input \mbox{--- with} different element types (e.g., forwarding devices, links, \mbox{paths) ---}, and custom message-passing definitions divided in multiple stages~\cite{rusek2019unveiling, list-papers}, where in each stage a specific set of elements (e.g., paths) share their hidden states with some other elements (e.g., devices).

\begin{figure}[!t]
    \centering
    \vspace{0.1cm}
    \includegraphics[width=0.9\columnwidth]{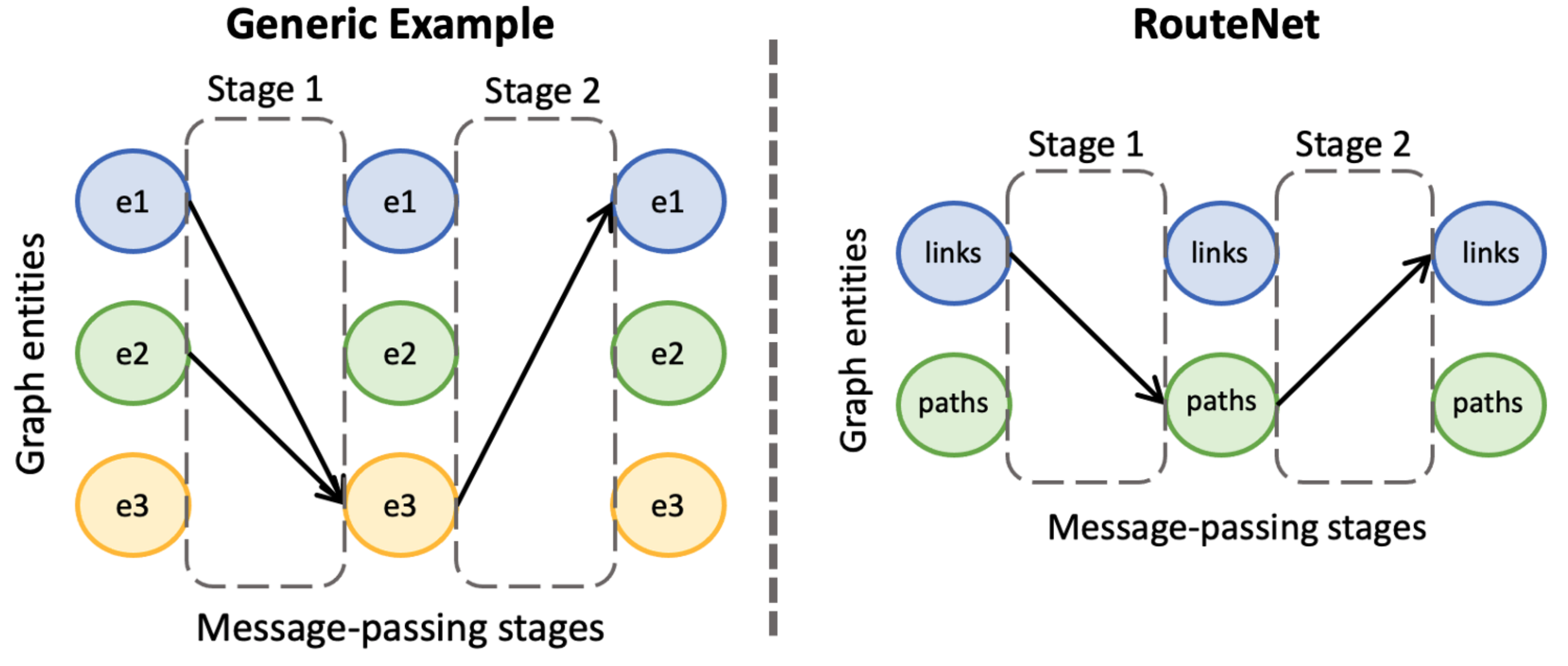}
    \caption{MSMP graphs --- Generic example and RouteNet~\cite{rusek2019unveiling}.}
    \label{fig:abstraction_graph}
\end{figure}
\setlength{\textfloatsep}{0.3cm}

Existing general-purpose GNN libraries offer high-level abstractions to simplify the implementation of custom GNN models. Among the most popular libraries, DGL~\cite{wang2019dgl} provides support for heterogeneous graphs, offering flexibility to define GNNs with differentiated message, aggregation, and update functions for each element type of input graphs, as well as combining components from well-known GNN architectures (e.g., Graph Convolutional Networks, Graph Attention Networks). Likewise, PyTorch Geometric~\cite{fey2019fast} provides pre-implemented GNN layers that can be combined to create complex GNN architectures. However, the higher-level abstractions of these libraries do not provide direct support for implementing multi-stage message passing definitions, with flexibility to define message exchanges between different element types at different execution stages, which is a common architectural framework among state-of-the-art GNN models for networking~\cite{rusek2019unveiling,list-papers}. This eventually requires the use of lower-level abstractions to implement these types of models.

In light of the above, IGNNITION introduces a novel high-level abstraction called the \textit{Multi-Stage Message Passing graph (MSMP graph)}, which is carefully designed to facilitate the implementation of GNN models applied to networking. With MSMP graphs, users can define their custom GNN architectures through a visual graph representation, abstracting them from the complex mathematical formulation behind their designs.Particularly, a GNN can be defined by a set of element types (hereafter, \textit{graph entities}) and how they relate to each other in a sequential order, allowing the definition of multi-stage message passing schemes in a natural and intuitive way. 

Figure~\ref{fig:abstraction_graph} (left) illustrates a generic example of an MSMP graph with three graph entities ($e1$, $e2$ and $e3$). In this scheme, two message-passing stages can be differentiated: First, elements of type $e1$ and $e2$ send their hidden states to their neighbors of type $e3$, according to their connections in the input graph. Then, in a second stage $e3$ elements share their states with their linked elements of type $e1$ and $e2$. Likewise, Figure~\ref{fig:abstraction_graph} (right) depicts the MSMP graph of RouteNet~\cite{rusek2019unveiling}. In this case, the GNN model considers two graph entities (\textit{links} and \textit{paths}), and implements a two-stage message-passing scheme that explicitly represent the circular dependencies between these two entities. The implementation of the \textit{Message}, \textit{Aggregation}, \textit{Update}, and \textit{Readout} functions is later defined by the user via an intuitive model description file in YAML, as described in the next section of this article. An MSMP graph description eventually describes a message-passing iteration of the GNN. Then, IGNNITION automatically unrolls this definition a number of iterations $T$ defined by the user.

At the same time, the MSMP graph abstraction offers a flexible modular design, which supports a wide range of existing GNN variants (e.g., Message-Passing NNs, Graph Convolutional Networks, Gated NNs, Graph Attention Networks, Graph Recurrent Networks)~\cite{ZHOU202057}, as well as custom combinations with individual components of them (e.g., message, aggregation, update, normalization). To the best of our knowledge, this abstraction supports all the state-of-the-art GNN models applied to networking at the time of this writing.

\vspace{-0.2cm}
\section*{IGNNITION Architecture} \label{Sec: Framework implementation} 
This section provides an overview of IGNNITION. In particular, we describe below the four main building blocks that shape this framework:
\begin{enumerate}
\item Model Description Interface
\item Dataset Interface
\item Core Engine
\item Debugging Assistant
\end{enumerate}
\vspace{-0.4cm}

\subsection*{Model Description Interface} \label{sec:Model-description}

The first step to build a GNN with IGNNITION is to describe the model architecture using the MSMP graph abstraction, introduced in the previous section. This can be done through a codeless interface, by filling a short YAML file. More specifically, this file contains the following information:

\subsubsection*{Entities Definition} \label{Entities definition}
First, the user defines the different entities (i.e., element types) to consider in the networking scenario and, consequently, in the corresponding MSMP graph. In the networking context, an entity represents a set of network elements, which can be physical (e.g., routers, links), or logical (e.g., paths, virtual network functions). As an example, Fig.~\ref{fig:implementation_example} shows the definition of an entity in the YAML model description file. There, the user indicates the entity name (e.g., \textit{path}), the size of the hidden state vectors (e.g., 32 elements), and the method and features used to initialize these vectors. The initialization can be defined as a flexible pipeline of operations that supports elaborate methods used in state-of-the-art models (e.g., normalization, initial feature embedding). Feature names (e.g., \textit{traffic}) are used as unique identifiers to then feed the model with any input dataset, as shown later in this section. 

\lstset{ %
  backgroundcolor=\color{white},
  basicstyle=\fontsize{8}{8}\ttfamily,
  breakatwhitespace=false,
  breaklines=true,
  captionpos=b,
  commentstyle=\color{commentsColor}\textit,
  escapeinside={\%*}{*)},
  extendedchars=true,
  frame=tb,
  keepspaces=true,
  keywordstyle=\color{black}\textbf,
  language=c,
  otherkeywords={*,name, entities, input,initial_state, state_dimension, type, source_entity, destination_entity, message, aggregation, update, nn_, adj_list, units, kernel_regularizer, activation, rate, architecture, source_entities, source_, stage_, stages, num_iterations, message_passing, message_passings  },          
  rulecolor=\color{black},
  showstringspaces=false,
  showtabs=false,
  stepnumber=1,
  tabsize=2,
  title=\lstname,
  columns=nice_alignment
}

\subsubsection*{Message Passing} \label{Message passing}
After defining the entities, the user completes the MSMP graph by describing the message-passing operations in the GNN model. These are based on the relationships between the graph entities previously defined. To do so, the user defines a single message-passing iteration, which can be divided into multiple stages. In each stage, a set of entities share their hidden states with other specific entities, thus offering flexibility to define the source and destination entities participating in each message-passing stage.

For example, in the generic MSMP graph of Fig.~\ref{fig:abstraction_graph} (left), the message passing is divided in two stages. In the first one, the following message exchanges are executed: $e1 {\rightarrow} e3$, and $e2 {\rightarrow} e3$; while the second stage performs the reverse message-passing operations: $e3 {\rightarrow} e1$, and $e3 {\rightarrow} e2$. 

\begin{figure}[!t]
    \centering
    \vspace{0.1cm}
\begin{lstlisting}
    # Entity definition
    entities:
    - name: path
      state_dimension: 32
      initial_state:
      - type: build_state
        input: [traffic]
    
    ...
    
    # Message-passing definition
    message_passing:
      num_iterations: N
      stages:
        # Stage 1
        - stage_message_passings:
          - destination_entity: path
            source_entities:
              - name: link
                message:
                  - type: direct_assignment
            aggregation:
              - type: ordered
            update:
              type: neural_network
              nn_name: recurrent1
\end{lstlisting}
\vspace{-0.8cm}
\caption{Example of a YAML model description file.}
\label{fig:implementation_example}
\end{figure}
\setlength{\textfloatsep}{0.2cm}

In the model description file, this can be defined as a series of YAML objects describing the several stages. For each stage, the user defines the names of the \textit{source} and \textit{destination} entities, and the \textit{Message}, \textit{Aggregation} and \textit{Update} functions to be applied. IGNNITION provides great flexibility to implement these functions, using any type of NN (e.g., multi-layer perceptron, recurrent, convolutional), or other operations such as element-wise sum, average, max, min, product, or even direct variable assignments, which are often used in state-of-the-art GNN models for networking~\cite{rusek2019unveiling, list-papers}. A full description of all the options available can be found at~\cite{ignnition_docu}. As an example, Fig.~\ref{fig:implementation_example} shows the definition of \textit{Stage 1} in the message passing of RouteNet~\cite{rusek2019unveiling} (see the corresponding MSMP graph in  Fig.~\ref{fig:abstraction_graph}-right). Note that, in case of using NNs (e.g., in the \textit{Update} function), the user can define a name (e.g., \textit{recurrent\_1}) that is then used to describe the NN details in a separate section of the file, as detailed later in this section.

\vspace{0.2cm}
\subsubsection*{Readout} \label{Readout}
After the message passing, the user defines the \textit{Readout} function, producing either per-node or global graph-level outputs. In IGNNITION, the \textit{Readout} is defined via a flexible pipeline of instructions that may combine \textit{pooling} operations, \textit{neural networks}, and other less common operations used in state-of-the-art GNN models applied to networking. \textit{Pooling} operations (e.g., element-wise sum, mean) transform a variable-size set of nodes' hidden states into a single fixed-size embedding vector, which is often used to make global graph-level predictions. Likewise, \textit{neural network} operations take as input nodes' hidden states \mbox{--- or} the embeddings resulting from previous \mbox{operations ---} and pass them through a NN.
More information about the \textit{Readout} definition can be found at~\cite{ignnition_docu}.

\vspace{0.1cm}
\subsubsection*{Neural Networks Definition} \label{Sec: Defining neural networks}
Lastly, the user can define all the neural networks previously referenced in a separate section (e.g., \textit{Message}, \textit{Update}, \textit{Readout} NNs). For this purpose, IGNNITION supports all the native functions of the well-known Keras library~\cite{chollet2015keras}. Thus, to define for instance a multilayer perceptron, the user should specify in the YAML file at least the fields required by Keras, which are the number and type of layers and some basic parameters (e.g., neurons per layer). For other undefined parameters, IGNNITION uses the Keras default values, thus considerably simplifying the definition of the NN. This YAML-based interface provides flexibility to implement any NN as long as it is supported by Keras, which is widely considered as one of the most complete low-level NN programming libraries.

\vspace{-0.2cm}
\subsection*{Dataset Interface} \label{sec:dataset}

The \textit{Dataset Interface} is intended to decouple the GNN model description from the input data. Similar to other GNN libraries, such as DGL~\cite{wang2019dgl} or PyTorch Geometric~\cite{fey2019fast}, IGNNITION enables to directly process data with the well-known NetworkX library. This Python library already implements a plethora of functions that automatize the definition of graphs from datasets, which enables to easily feed the GNN model with datasets from different sources and in various formats (e.g., NetFlow/IPFIX measurement reports, router configuration logs).

\vspace{-0.3cm}
\subsection*{Core Engine} \label{sec:IGNNITION}

This module implements the main logic behind IGNNITION. Once the model has been designed and the datasets have been properly formatted --- as explained above --- the user can call different functionalities from the \textit{Core Engine}:

\begin{itemize}
    \item Generate a TensorFlow implementation of the GNN
    \item Run the training/validation of the model
    \item Make predictions with a trained model
    \item Generate visual representations of the model (for debugging purposes)
\end{itemize}

The \textit{Core Engine} internally implements a generic definition of the MSMP graph abstraction, thus covering the broad spectrum of GNN architectures supported by such abstraction. To generate the model implementation, the \textit{Core Engine} proceeds in a similar way to traditional compilers, first performing a lexical analysis, then a syntactical analysis, and finally a semantic analysis. These three steps enable to detect unexpected structures in the model definition (in YAML files) and trigger numerous error-checking messages to help users debug their models.

After validating all the input information, IGNNITION automatically generates the GNN model implementation directly in TensorFlow. This enables to work internally with the efficient computational graph produced by this ML library. As a result, IGNNITION implementations have comparable performance to models directly coded in TensorFlow, as shown later in the evaluation of this article.

\vspace{-0.3cm}
\subsection*{Debugging Assistant} \label{sec:debugging}
One of the biggest challenges when designing a GNN model with traditional ML libraries is to identify potential bugs and fix them. Typically, having a clear picture of the resulting GNN model is a cumbersome and time-consuming task. For this reason, IGNNITION incorporates an advanced debugging system that assists the user in different ways.

First, it automatically produces an interactive visual representation of the internal GNN architecture. To this end, it relies on \textit{Tensorboard} --- a TensorFlow-based visualization \mbox{toolkit ---} allowing the user to dynamically visualize its GNN model at different levels of granularity. Standard visualizations of TensorBoard show the resulting computational graph of the model. However, it is difficult to identify the different building blocks of the GNN from the tensor operations shown in this graph (e.g., message, aggregation, update functions, message-passing iterations). IGNNITION implements a layer on top of this computational graph (using scope variables) that differentiates the main building blocks of the GNN at different levels of hierarchy, which makes it easier to identify the internal functions. For example, after implementing RouteNet~\cite{rusek2019unveiling} with IGNNITION the user can navigate through the resulting TensorBoard graph and easily identify the two message-passing stages of this model (see Fig.~\ref{fig:abstraction_graph}-right): \textit{links$\rightarrow$paths}, and \textit{paths$\rightarrow$links}. Moreover, by zooming in on the visualization, the user can easily find the internal functions and how they are implemented (e.g., \textit{Message}, \textit{Aggregation}, \textit{Update}).

Additionally, the debugging assistant incorporates several advanced error-checking mechanisms to ensure the correct definition of the model and to provide guidance to fix badly defined fields in YAML files (e.g., wrong NN descriptions, entities definitions, missing features in the datasets).

\vspace{-0.1cm}
\section*{Case Studies} \label{Sec: GNN design use cases}
GNNs have been successfully applied to a plethora of networking use cases, such as routing optimization~\cite{rusek2019unveiling, geyer2018learning}, virtual network function placement~\cite{quang2019deep}, resource allocation in wireless networks~\cite{eisen2020optimal}, job scheduling in data center networks~\cite{mao2019learning}, MPLS configuration, or Multipath TCP. A more comprehensive list of relevant GNN-based applications for communication networks can be found at~\cite{list-papers}.

\begin{figure}[!t]
    \centering
    \vspace{0.1cm}
    \includegraphics[width=\columnwidth]{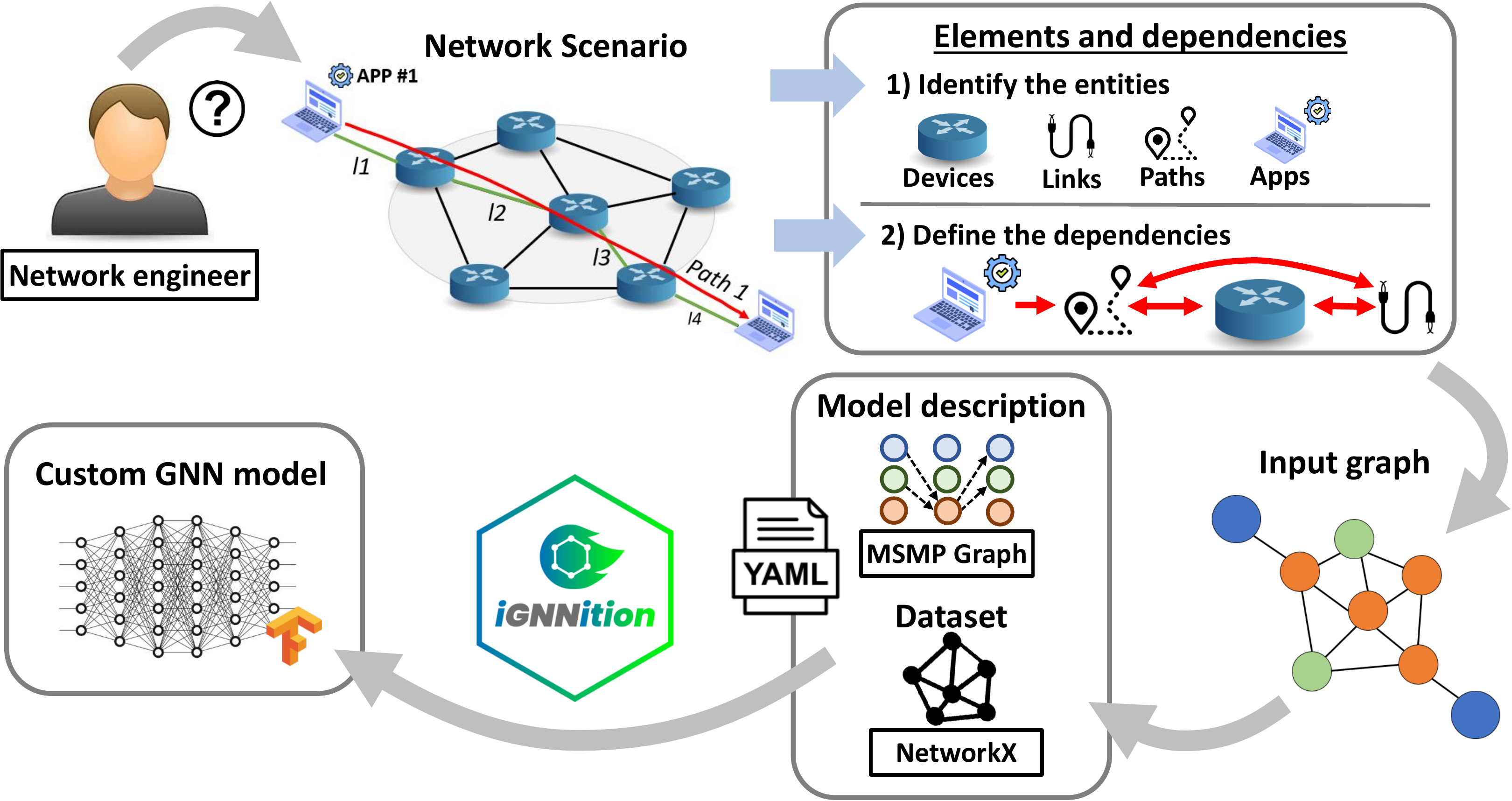}
    \caption{General workflow to produce GNN models for networking with IGNNITION.}
    \label{fig:case-studies}
\end{figure}

Figure~\ref{fig:case-studies} illustrates the general workflow for producing with IGNNITION a GNN model adapted to a particular network problem. First, the user --- e.g., a network engineer --- needs to identify the main elements involved in the networking problem to be addressed (e.g., devices, links, apps, paths) and, according to the purpose of the GNN model (e.g., Traffic Engineering), define potential dependencies between these different elements. In the networking context, we can often find circular dependencies between different elements that can be naturally encoded in GNNs as a sequence of message-passing stages. These elements and dependencies shape the input graphs of the GNN, thus defining the structure of the samples in the train, validate and/or test datasets used for the model. Afterward, the user can define the architecture of the GNN. In IGNNITION, this can be done by filling the YAML model description file, leveraging the novel MSMP graph abstraction. Lastly, IGNNITION generates an efficient implementation of the model in TensorFlow that can be trained with the generated datasets.

As an example, we briefly describe below how we can implement with IGNNITION two state-of-the-art GNN models that introduce some non-standard architectural patterns.

\setlength{\textfloatsep}{0.2cm}

\vspace{-0.3cm}
\subsection*{RouteNet}
RouteNet~\cite{rusek2019unveiling} was proposed as a network modeling tool that predicts path-level performance metrics (e.g., delay, jitter) given a network state snapshot as input, which is defined by: a network topology, a routing configuration, and a traffic matrix. This GNN is then used to optimize the routing configuration in networks, by combining the model with an optimizer. To this end, RouteNet considers input graphs with two different entities (\textit{links} and \textit{paths}) that have circular dependencies between them, and encodes its input parameters as features within the initial hidden states of these elements. As a result, this GNN model implements a two-stage message-passing scheme combining the hidden states of these entities, which can be naturally defined by an MSMP graph (see Fig.~\ref{fig:abstraction_graph}-right).

\vspace{-0.25cm}
\subsection*{Graph-Query Neural Network (GQNN)}
The GQNN model~\cite{geyer2018learning} addresses a different problem: supervised learning of traditional routing protocols with GNN, such as shortest path or max-min routing. To this end, this GNN model uses a novel architecture with two graph entities: \textit{routers} and \textit{interfaces}, being the latter the several network interfaces of each router in the network. This model considers a single-stage message-passing scheme where routers and interfaces share their hidden states simultaneously. As output, it determines whether the interfaces are used to transmit traffic or not (i.e., [0,1]), which eventually defines the routing configuration of the network. GQNN introduces a particularity in the \textit{Readout} definition, which uses a non-standard operation pipeline with an element-wise product and then a NN to produce the final prediction. With IGNNTION, this can be easily defined in the YAML model description file as a pipeline with a \textit{product} operation and then a NN function.

\rm{For more information on how to implement a GNN prototype with IGNNITION, we refer the reader to~\cite{ignnition_docu}. There, the user can find an easy-to-follow quick-start tutorial, where a basic GNN model is built to compute the shortest path routing configuration given a network topology with weights on links. Also, IGNNITION includes a library with state-of-the-art GNN models applied to different networking use cases, which is under continuous development by the community of this open-source project.}

\vspace{-0.25cm}
\section*{Evaluation: Accuracy and Cost}

As mentioned earlier, IGNNITION internally implements GNN models in TensorFlow, thus leveraging the efficient computational graphs produced by this ML library. This section aims to showcase this aspect by making a direct  comparison of IGNNITION implementations with respect to implementations directly coded in TensorFlow. As an example, we consider the two state-of-the-art GNN models described in the previous section (RouteNet and GQNN).

For the evaluation, we use the implementations of these models in IGNNITION --- publicly available at~\cite{ignnition_docu} --- and, as a reference, their native implementations in TensorFlow, and reproduce some of the experiments made in their corresponding papers~\cite{rusek2019unveiling,geyer2018learning}. Note that IGNNITION does not provide additional compiling optimizations. Instead, in the evaluation we reproduce the same scheme used in the original TensorFlow implementation. We use the datasets of the original evaluations of RouteNet, and GQNN (with \textit{300,000}, and \textit{40,000} samples respectively), where 80 percent of the samples are randomly selected for training and 20 percent for evaluation.

We first evaluate the accuracy achieved by the models, to check that both implementations achieve the same results under equal training conditions. Our experimental results show that the IGNNITION implementation of RouteNet yields a Mean Relative Error of $2.62$ percent, and the implementation of GQNN an accuracy of $98.07$ percent. Both numbers are in line with the results obtained by the original TensorFlow implementations.

\begin{figure}[!t]
    \centering
    \vspace{-0.4cm}
    \includegraphics[width=0.9\columnwidth]{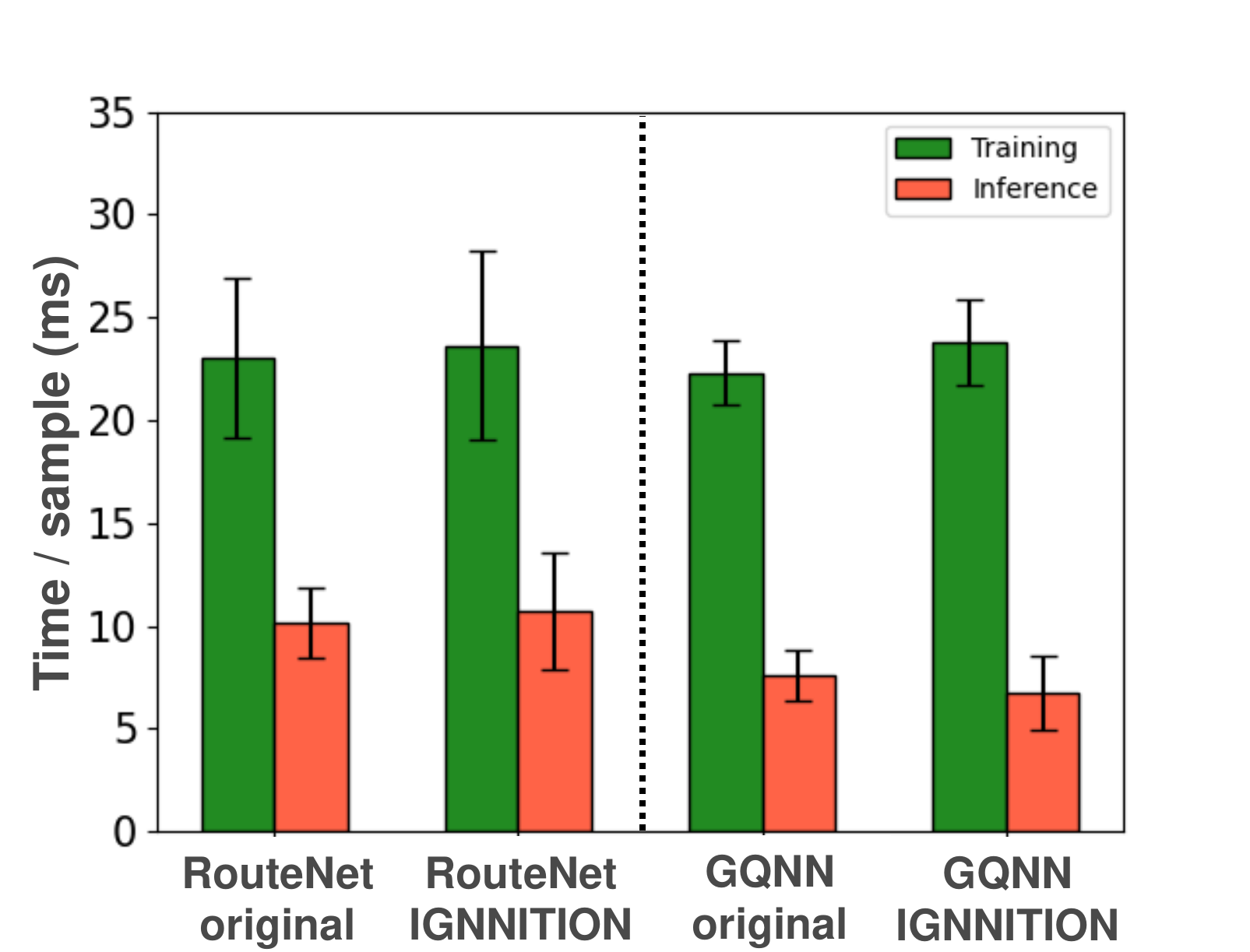}
    \caption{Evaluation of the execution time of IGNNNITION vs. native TensorFlow implementations (training and inference).}
    \label{fig:comparison_execution}
\end{figure}

Likewise, we evaluate the execution cost of both types of implementations. Figure~\ref{fig:comparison_execution} depicts the average execution time per sample during training and inference. In line with the previous results, we can observe that the execution times of IGNNITION implementations are equivalent to those of the original models implemented in TensorFlow. Note that all these experiments were made in a controlled testbed, considering the same computing resources for all cases.

These results reveal the capability of IGNNITION to produce efficient GNN models, while offering substantial time savings for non-ML experts to implement them. Moreover, users can leverage additional functionalities such as the visual representations of the debugging system and the advanced \mbox{error-checking} mechanisms to easily fix potential bugs.

\section*{Related Work} \label{Sec: Related work}
Several GNN libraries have emerged in recent years, motivated by the outstanding applications of GNNs to a plethora of relevant real-world problems. Among the most popular open-source libraries, we can find: DGL~\cite{wang2019dgl}, PyTorch Geometric~\cite{fey2019fast}, or Graph Nets~\cite{battaglia2018relational}. Also, NeuGraph~\cite{ma2019neugraph} has recently become a popular GNN library that focuses on parallelizing the execution of GNN models, although it is not open source. These libraries provide high-level abstractions to code standard GNN layers (e.g., Graph Convolutional Networks, Graph Attention Networks) that can be concatenated. However, these higher-level abstractions do not provide direct support for implementing multi-stage message passing iterations, where in each stage specific graph entities share their hidden states with some other entities, which is a common architectural pattern for modeling circular dependencies in GNN models for networking~\cite{rusek2019unveiling,list-papers}. As an example, DGL provides several built-in functionalities to deal with heterogeneous graphs. Nevertheless, these functionalities do not handle directly the implementation of message-passing definitions where different graph entities share their hidden states sequentially. As a result, it is eventually needed to use lower-level abstractions to implement these types of models. Lastly, some other popular libraries, such as GraphGym~\cite{you2020design} or Spektral~\cite{grattarola2020graph}, provide higher-level abstractions compared to the previous libraries. However, these libraries do not support many non-standard GNN architectures used in networking (e.g.,~\cite{rusek2019unveiling, geyer2018learning}).

In contrast, IGNNITION provides a \textit{codeless interface} that users can leverage to easily implement GNN prototypes. This interface is based on the novel MSMP graph abstraction proposed along with this framework, which is inspired by the architectural patterns commonly used in GNN models for networking. In particular, this abstraction naturally introduces the definition of non-standard message-passing architectures considering heterogeneous graphs with sequential dependencies between different sets of elements.

\section*{Conclusion} \label{Sec: Conclusion}
In this article, we have introduced IGNNITION, an open-source framework based on TensorFlow that enables fast prototyping of \textit{Graph Neural Networks} for networking applications. IGNNITION works over a novel high-level abstraction called the \textit{Multi-Stage Message Passing graph (MSMP graph)}, which isolates users from the complex mathematical formulation and implementation in traditional ML libraries (e.g., TensorFlow, PyTorch). MSMP graphs are flexible enough to support state-of-the-art GNN architectures with non-standard message-passing strategies. We implemented with IGNNITION two state-of-the-art GNNs for networking, to showcase the versatility of this framework. Likewise, we validated that the performance of the GNN models produced by IGNNITION is equivalent to that of native TensorFlow implementations, as efficiency is a must for networking \mbox{applications}.

\section*{Acknowledgment}
This open-source project has received funding from the European Union’s Horizon 2020 research and innovation programme within the framework of the NGI-POINTER Project funded under grant agreement No. 871528. This article reflects only the author's view; the European Commission is not responsible for any use that may be made of the information it contains. The work was also supported by the Spanish MINECO under contract TEC2017-90034-C2-1-R \mbox{(ALLIANCE)} and the Catalan Institution for Research and Advanced Studies (ICREA).

\bibliographystyle{IEEEtran}
\bibliography{references}

\section*{Biography}
\label{sec:auth}

\textbf{David Pujol-Perich} is an MSc student in advanced computing from the Universitat Politècnica de Catalunya (2020-2022), where he also received his BSc in Computer Science (2020). In 2019-2020, he did an academic exchange in ETH Zurich. He is currently a machine learning researcher at the Barcelona Neural Networking Center (BNN-UPC).

\textbf{José Suárez-Varela} received his PhD in Computer Science from the Universitat Politécnica de Catalunya (2020). He is currently a postdoctoral researcher at the Barcelona Neural Networking Center (BNN-UPC), and co-Principal Investigator of the EU-funded project IGNNITION (NGI Pointer). His main research interests are in Machine Learning applied to networking.

\textbf{Miquel Ferriol-Galmés} received his BSc in Computer science (2018) and MSc in Data Science (2020) from the Universitat Politècnica de Catalunya. He is currently pursuing a PhD at the Barcelona Neural Networking center (BNN-UPC). His main research interests are in the application of Graph Neural Networks to computer networks.

\textbf{Shihan Xiao} received a B.Eng. degree in electronic and information engineering from the Beijing University of Posts and Telecommunications, in 2012, and a PhD degree from the Department of Computer Science and Technology, Tsinghua University. He is currently a senior engineer with Huawei 2012 NetLab.

\textbf{Bo Wu} received his Bachelor’s degree from the School of Software at Shandong University, China, in 2014, and his PhD degree from the Department of Computer Science and Technology at Tsinghua University in 2019. Currently, he works in the Network Technology Laboratory at Huawei Technologies.

\textbf{Albert Cabellos-Aparicio} is an assistant professor at Universitat Politècnica de Catalunya (UPC), where he obtained his PhD in computer science in 2008. He is director of the Barcelona Neural Networking Center (BNN-UPC) and scientific director of the NaNoNetworking Center in Catalunya.

\textbf{Pere Barlet-Ros} is an associate professor at Universitat Politècnica de Catalunya (UPC) and scientific director of the Barcelona Neural Networking Center (BNN-UPC). From 2013 to 2018, he was co-founder and chairman of the machine learning startup Talaia Networks, which was acquired by Auvik Networks in 2018.

\end{document}